\documentclass{article}

\usepackage[preprint]{neurips_2024}

\usepackage{amsmath}
\usepackage{graphicx}
\usepackage{svg}

\usepackage[utf8]{inputenc} 
\usepackage[T1]{fontenc}    
\usepackage{hyperref}       
\usepackage{url}            
\usepackage{booktabs}       
\usepackage{amsfonts}       
\usepackage{amsmath}
\usepackage{nicefrac}       
\usepackage{microtype}      
\usepackage{xcolor}         
\usepackage{subcaption}     

\title{HRM-Agent: Training a recurrent reasoning model in dynamic environments using reinforcement learning}

\author{%
  Long H Dang \\
  Melbourne, Australia \\
  \texttt{hoanglongdang2001@gmail.com} \\
  \And
  David Rawlinson \\
  Cerenaut AI \\
  Melbourne, Australia \\
  \texttt{dave@cerenaut.ai} \\
}
\DeclareMathOperator{\MSE}{MSE}

\begin{document}

\maketitle

\begin{abstract}
The Hierarchical Reasoning Model (HRM) has impressive reasoning abilities given its small size, but has only been applied to supervised, static, fully-observable problems. One of HRM's strengths is its ability to adapt its computational effort to the difficulty of the problem. However, in its current form it cannot integrate and reuse computation from previous time-steps if the problem is dynamic, uncertain or partially observable, or be applied where the ``correct'' action is undefined, characteristics of many real-world problems.

This paper presents HRM-Agent, a variant of HRM trained using only reinforcement learning. We show that HRM can learn to navigate to goals in dynamic and uncertain maze environments. Recent work suggests that HRM's reasoning abilities stem from its recurrent inference process. We explore the dynamics of the recurrent inference process and find evidence that it is successfully reusing computation from earlier environment time-steps.
\end{abstract}

\section{Introduction}

\subsection{Reasoning models}
Most Large Language Models (LLMs) sequentially generate a stream of output tokens as part of a forward inference process of pre-determined computational complexity. The inference process does not intrinsically adapt to the complexity of the task, leading to inefficiency. Unreliable performance is often reported, especially in tasks (including maze navigation) where the model must jointly review all its outputs to validate an appropriate response \citep{jolicoeurmartineau2025less}.  

Several techniques have been proposed to remedy this weakness, including ``Chain of Thought'' (CoT) \citep{wei2022chain} and Tree-of-Thought (ToT) exploration \citep{yao2023tree}, in which the model explicitly generates intermediate steps before outputting a final decision; ``Thinking tokens'' \citep{herel2024thinking}, which provide additional inference steps; scaling test-time compute \citep{snell2024testtimecompute}, and ``agentic'' architectures such as the ``AI Scientist'' \citep{lu2024aiscientist}. However, all these strategies ultimately exploit \emph{human} insight and ingenuity to devise ways to coax better results from broadly unchanged models. Most current agentic systems require \emph{human} thought to design a process to decompose broad problems into smaller sub-problems the model will likely find more manageable, or where it will perform more consistently. This seems inadvisable given that one of the key lessons from deep learning is that learned models tend to outperform human-defined problem decompositions \citep{lecun2015deep}. 

\cite{wang2025hierarchicalreasoningmodel} recently proposed the Hierarchical Hierarchical Reasoning Model (HRM) which demonstrated data-efficient reasoning in a relatively compact model. Importantly, given the complaint about LLMs above, HRM has a recurrent inference process which can adapt to the complexity of the current task. The HRM paper includes a demonstration of automatic, inference-time scaling of computational complexity using the Adaptive Computation Time (ACT) component, which determines the number of recurrent iterations to perform.

HRM employs dual recurrent modules - a ``high-level'', slow-updating module $H$, and a ``low-level'', fast-updating module $L$. The different levels of abstraction in $H$ and $L$ are intended to emerge from convergence dynamics and are not explicitly enforced.

Both models work together, initially in a recurrent mode to convergence, and then in a single forward pass without explicit supervision of intermediate steps to produce a complete solution to a fully-observable, static problem. In their paper, this architecture is shown to solve complex reasoning tasks such as extreme Sudoku, maze path planning, and tasks from the Abstraction and Reasoning Corpus (ARC) \citep{chollet2019measureintelligence} with orders of magnitude fewer parameters ($\sim27$M) and training examples ($\sim1$k) than traditional deep reasoning models.

HRM's reported combination of performance and efficiency is interesting because much larger reasoning models appear to plateau in performance despite scaling the available computational resources \citep{illusion-of-thinking}, suggesting a fundamental limitation.

Since publication of the HRM paper, other authors have explored derivative models. In a blog post by the ARC-prize foundation \citep{arcprize2025hrm} it was reported that the largest contributor to model performance was the recurrent process. More recently, \cite{jolicoeurmartineau2025less} published the Tiny Recursive Model (TRM). Via a series of ablations and variants, the TRM paper reports that the recurrent (recursive) forward process is key to its reasoning abilities. The TRM combines the $H$ and $L$ modules but retains the alternating pivot between two latent vectors during recurrent convergence. 

\subsection{Reasoning in dynamic, uncertain or partially observable environments}
If HRM can unlock improved reasoning abilities in static and fully observable tasks, can these benefits be realized in dynamic, uncertain or partially observable environments?

This paper aims to explore whether the reasoning abilities of HRM can be used in the context of an agent which is trained using only reinforcement learning, without supervised losses. This would enable HRM to be applied to problems which cannot be tackled by simply by analysis of the initial problem state. In these problems, an agent must incrementally develop a solution via a series of actions, and continually adapt its solution to changes in the environment.

Reinforcement learning (RL) has been successfully applied in various domains, from Atari gameplay \citep{mnih2015dqn} to robotic control \citep{kober2013rlrobotics}. RL is appropriate when it is impractical or simply not meaningful to define a correct response. Instead, models maximize \emph{qualitative} feedback as given as rewards. Rewards may be sparse, meaning that most states provide no feedback at all. This makes RL suitable for tasks which require incremental solution construction.

``Reasoning'' in a RL context can be expressed in many ways, including long horizon path planning \citep{sutton1999options}, multi-step instruction following \citep{chevalier2023minigrid, chevalier2018babyai}, logical reasoning about entities \citep{zambaldi2019relational}, discovery of causal, hierarchical sequences (e.g. making and using tools) \citep{hafner2021crafter}, arithmetic reasoning \citep{saxton2019mathematics}, theory of mind \citep{bard2020hanabi}, multi-agent planning \citep{vinyals2017starcraft} and abstract pattern generalization \citep{cobbe2019leveraging}. In this paper we have focused on the first of these, path planning.



Many real-world RL problems are dynamic, uncertain and partially observable (PO). For example, a mobile robot must move around to sense different locations in the environment. Partially Observable Markov Decision Processes (POMDPs) \citep{kaelbling1998pomdp} capture settings where the agent cannot directly observe the full environment state, requiring it to maintain an internal belief over possible states. In dynamic and uncertain environments, where transition dynamics and relevant observations may change unpredictably, agents must continually integrate new information, while discarding outdated knowledge. Solution construction is usually incremental and cannot be determined in a single observation–action cycle regardless of the initial reasoning effort or quality.

RL problems can be fully-observable but still dynamic and uncertain. In a Markov Game such as Chess or Go the state of the board is observable but the actions of the opposing player are unpredictable. AlphaGo \citep{silver2016alphago} used deep reinforcement learning in combination with Monte Carlo Tree Search to demonstrate reasoning and planning over long time horizons.

Q-learning \citep{watkins1992qlearning} is a widely studied off-policy approach to RL.
This makes it well-suited to environments where exploration is costly, as past trajectories can be reused for policy improvement. Extensions such as Deep Q-Networks (DQNs) \citep{mnih2015dqn}, Double DQNs \citep{vanhasselt2016double}, and distributional Q-learning \citep{bellemare2017distributional} have extended the approach to high-dimensional and stochastic settings.

However, vanilla Q-learning agents struggle when the mapping from historical observations to optimal actions involves hierarchical, long-horizon reasoning. \cite{hausknecht2015drqn} applied a ``Deep Recurrent Q Network'' (DQRN) to partially observable domains by replacing the fully connected layer before the output with an Long-Short-Term-Memory (LSTM) layer \citep{hochreiter1997long}. This allows the Q-function to condition action values on a learned embedding of recent observation history. DRQN demonstrated improved performance in Atari games,
and showed that explicit recurrence can help an agent maintain and update an internal belief state. This insight suggests that when adapting HRM to a dynamic and/or PO context, recurrent inputs should be maintained between environment steps to promote consistent evaluation of intended plans.

RL is also widely used to fine-tune LLMs - notably reinforcement learning from human feedback (RLHF) \citep{christiano2017rlhf} and reinforcement learning from AI feedback (RLAIF) \citep{bai2022constitutional}. Fine-tuning often aims to improve reasoning quality, factual accuracy, and task adherence. In more agent-like contexts, RL-based fine-tuning can optimize the model for goal-directed multi-step interaction, adjusting its internal reasoning process to maximize downstream success metrics rather than immediate predictive likelihood.

However, LLM-based reasoning approaches face several challenges in partially observable and dynamic environments. CoT and ToT methods often require repeated autoregressive rollouts, leading to high inference latency, and they typically assume that all relevant information is accessible at reasoning time. In contrast, RL agents in POMDPs gather information incrementally through exploration before reasoning can converge on a solution, suggesting a need for architectures that interleave memory, reasoning, and action selection across multiple timesteps.



\section{Method}
We used reinforcement learning to train a variant of the HRM model to successfully navigate to goals in two fully-observable, dynamic and uncertain maze environments. 

\subsection{Environments}
The HRM paper \citep{wang2025hierarchicalreasoningmodel} demonstrated successful reasoning on three tasks - Sudoku, ARC-AGI and maze path-planning. Of these, maze path-planning seems most easily adapted to be a dynamic problem, by placing and removing obstacles. In a RL context, path-planning tasks are often framed as goal navigation - the Agent must navigate from a random start location to a random goal location. Successful and efficient navigation in maze environments implies that the model has learned a policy which can plan paths, because simple heuristic policies will not allow the agent to reach the goal quickly. We assume that path-planning is a reasonable demonstration of a form of reasoning.

In the original HRM paper, the supervised training regime provides rich, precise feedback to the model about every generated pixel or token. In a RL setting, feedback is typically both sparse and ambiguous, making learning more difficult. For this reason our initial experiments use smaller, simpler maze environments to validate that the HRM model can learn to reason in a RL setting.

The maze environments used in this paper are derived from the NetHack Learning Environment (NLE) \citep{kuttler2020nethack} via the MiniHack \citep{samvelyan2021minihack} and Gymnasium \citep{towers2024gymnasium} libraries. An episode begins with the agent being placed at a random start location and each observation includes the current state of the maze environment (agent, walls, doors, and location of the goal). Each episode ends after a defined maximum number of steps or when the agent reaches the goal. We have allowed four-way movement of the agent, in the cardinal directions. A small negative reward of $r==0.01$ is provided if the agent attempts to walk into an obstacle. The model configuration, loss, training parameters and reward function are identical for all environments.

\begin{figure}
  \centering
  \includegraphics[width=1.0\textwidth]{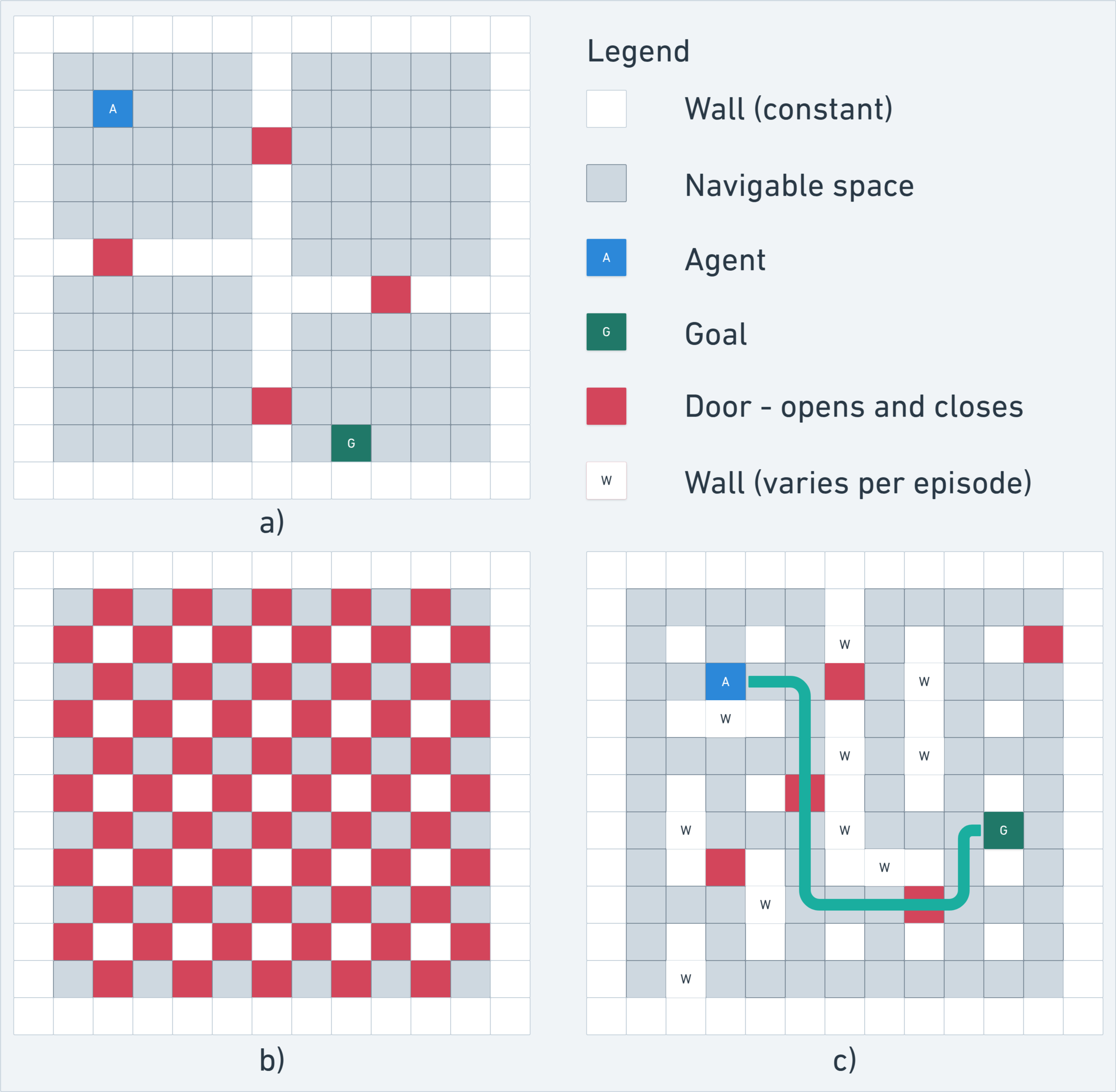}
  \caption{Maze-navigation tasks used to test the HRM-Agent model in the Nethack Learning Environment (NLE) with MiniHack. a) Four-rooms environment from \citep{sutton1999options} and \citep{rauber2019hindsight} with added doors (red). One of the four doors is closed at any given time. The closed door changes randomly during episodes. In each episode, the agent must navigate from a random start location to a random goal location. b) The random maze environment. This is a grid of fixed walls (white) and $N=10$ walls added randomly to some of the red squares before each episode. The random walls remain constant during the episode. In addition, a set of $M=5$ doors are enabled at the start of each episode. Each door randomly opens and closes independently during the episode. This creates a random corridor-type maze as shown in sub-figure (c), with viable paths changing as doors open and close. A* path-planning is used during maze generation to ensure that at least one viable path exists when all doors are open.}
  \label{fig:env}
\end{figure}

\subsubsection{Four-rooms environment}
The Four-rooms environment (figure \ref{fig:env}a) is a simple maze-navigation task used in several previous works including \cite{sutton1999options} and \cite{rauber2019hindsight}. A reward of 1 is provided if the agent reaches the goal location, otherwise the reward is zero. This environment is usually described as a sparse-rewards problem, because the narrow doorways make it improbable that the agent will randomly move between rooms. 

To make this environment dynamic and uncertain, we added doors to these doorways. One door is closed at all times. During each environment step, the closed door is randomly reassigned with probability $p=0.05$. This means that when a door is closed, it tends to stay closed for many steps, making an alternative path a better option. Due to the arrangement of the four rooms, there is always an alternative path to the goal when one door is closed (see figure \ref{fig:env}a). These dynamics mean that the agent's planned path to the goal may be disrupted at any moment, necessitating re-planning. Due to these characteristics, this environment is ideal for analysis of the model's recurrent state $z$.

\subsubsection{Dynamic, random maze environment}
This maze begins with a regular grid of permanent walls (figure \ref{fig:env}b), with $N=10$ fixed random walls and $M=5$ random doors (both of which can only be added at the red squares in figure \ref{fig:env}b). The random walls are added at the start of each episode and do not change during the episode. The random doors are added at the start of each episode, but independently, randomly open and close during each environment update. The probability of each individual door opening or closing is $p=0.05$. Note that in this environment, multiple doors can close simultaneously.

During maze generation we use A* path planning \citep{hart1968formal} to ensure that at least one path \emph{transiently} exists between the start and the goal, after the random walls have been added. However, it is possible that no path exists at times when certain doors are closed. In these instances, the agent must wait for a door to open to complete navigation. Figure \ref{fig:env}c depicts a typical generated maze.

The purpose of the dynamic random maze environment is to force the agent to learn to plan in response to a \emph{novel} maze rather than having simply learned the optimal policy for the single four-rooms environment. This is equivalent to asking the model to demonstrate a greater degree of generalization.

By having a very large number of possible mazes it reduces the probability that the Agent has learned the optimal path for all of them, and increases the probability that it instead learns a generalized policy to dynamically generate paths for any comparable maze. This capability is assumed to be a type of reasoning. Keeping the model parameter count small also makes learning a generalized planning capability more feasible than learning the optimal policy for all configurations of all the possible mazes.

\subsection{Model}
\begin{figure}
  \centering
  \includegraphics[width=1.0\textwidth]{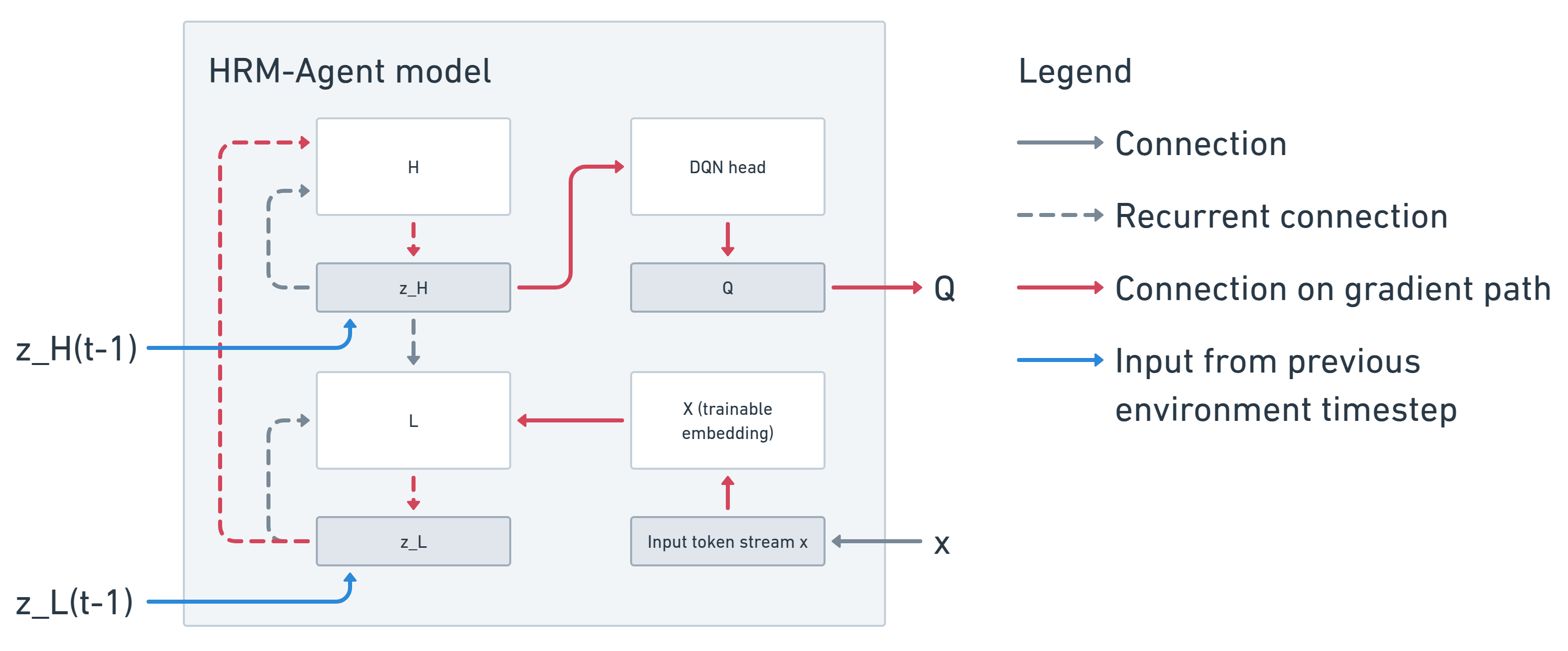}
  \caption{Architecture of the HRM-Agent model. The output head has been replaced with a fully-connected two-layer DQN head, which produces $Q$-values (expected rewards) for each action. Another change is the initialization of the recurrent state $z_L$ and $z_H$ from their final, converged values in the previous environment time-step. However, no gradients are propagated across environment time-steps. Like the original HRM paper, gradients are not accumulated during the recurrent forward process.}
  \label{fig:arch}
\end{figure}

We used the simplest possible model architecture to adapt the HRM model to be an RL agent. The model is trained entirely from random parameter initialization with reinforcement learning and does not receive any supervised loss, labels or target values. The HRM output head is replaced with a Deep-Q Network (DQN) head \citep{mnih2015dqn} and actions are selected using an Epsilon-Greedy exploration strategy. Therefore, the model is an ``off-policy'' one. During validation episodes, Epsilon is set to zero. Figure \ref{fig:arch} shows the modified HRM-Agent architecture. The model hidden size was reduced to 64 giving a total trainable parameter count of approx. 500,000. Other hyperparameter values used in our experiments are listed in the appendix.

The current model $\theta$ is trained to minimize the MSE loss (equation \ref{eqn:loss}) between the current model's predicted $Q$-values for the current environment step and the bootstrapped target values $y$. Target values are produced according to equation \ref{eqn:bellman} from the current environmental reward $r$ and the discounted, predicted reward of the next state $Q_{\theta^-}(s', a')$, generated from a target copy of the model $\theta^-$ whose weights are slowly updated from the current model. Note that $d$ is a terminal step indicator which excludes predicted rewards beyond the end of an episode:

\begin{equation}
Loss = \MSE(Q_{\theta}(s, a), y)
\label{eqn:loss}
\end{equation}

\begin{equation}
y = r + \gamma \, (1 - d) \, \max_{a'} Q_{\theta^-}(s', a')
\label{eqn:bellman}
\end{equation}

\subsubsection{Adaptive computation time}
The Adaptive Computation Time (ACT) component of the original HRM architecture has been disabled to facilitate analysis of the convergence of the recurrent state $z$, which has $z_L$ and $z_H$ elements for the ``Low'' and ``High'' levels of the hierarchy.

\subsubsection{Environment time and recurrent time}
The model now has \textbf{two} ``time-like'' dimensions: \textbf{environment time}, which are the steps within an episode, and \textbf{recurrent-time} which occurs within one environment time-step. This effectively allows the model unlimited thinking time without any change in the environment. 

In the original HRM model, the recurrent state $z$ is initialized to a random normal vector at the start of each forward pass. The latent vector $z$ is updated iteratively during the recurrent process, and a final forward and backward pass is used to produce $z$ for both output and gradient updates. No gradients are produced during the recurrent process.

In most environments it is likely that the contents of $z$ from the previous environment time-step are still relevant to the current environment time-step; in fact consistency is usually important for successful plan execution. 

In this paper, two HRM-Agent model variants were evaluated. In the ``carry Z'' variant, the initial value of $z$ is a random vector for the first step in an episode. In all other steps in an episode, the initial value of $z$ is the final value of $z$ from the \emph{previous} environment time step, ``carried forward''. No gradients are propagated through environment-time i.e. the initial value of $z$ is detached. No gradients are accumulated during the recurrent convergence process either, exactly as in the HRM paper. The idea of carrying forward a latent vector has been used before to good effect, such as in ``Training Large Language Models to Reason in a
Continuous Latent Space'' \citep{hao2024training} although in that work gradients are propagated with the latent vector. 

In the ``reset Z'' variant, the initial value of $z$ is always a random vector for all environment time-steps. This variant is intended as an ablation to contrast with the ``carry Z'' variant and explore whether the model can utilise ``planning'' from previous environment time-steps given changes to the environment and agent position.

\section{Results}
The reported experiments were designed to: 

\begin{itemize}
    \item validate that the HRM model could be adapted to learn to reason successfully in a RL setting so that it can be applied to dynamic or partially-observable problems, and 
    \item explore the dynamics of the recurrent ``reasoning'' process to verify that the model is able to utilise earlier computation as the environment changes. If the model cannot benefit from earlier computation, each time-step becomes an independent problem and the efficiency of the model in dynamic problems would be severely limited.
\end{itemize}

\subsection{Validation of concept}
\begin{figure*}[t!]
  \centering
    \begin{subfigure}[t]{0.49\textwidth}
      \centering
      \includegraphics[width=1.0\textwidth]{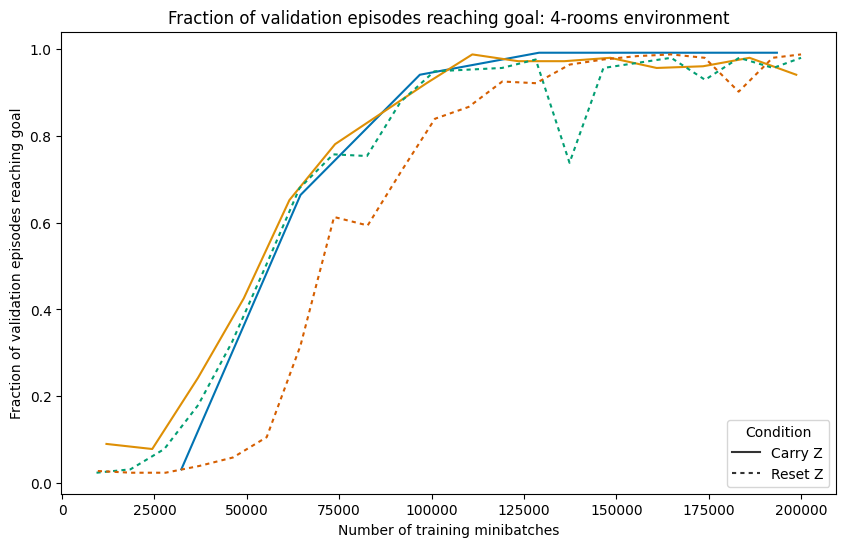}
      \caption{Four-rooms environment}
    \end{subfigure}
    \begin{subfigure}[t]{0.49\textwidth}
      \centering
      \includegraphics[width=1.0\textwidth]{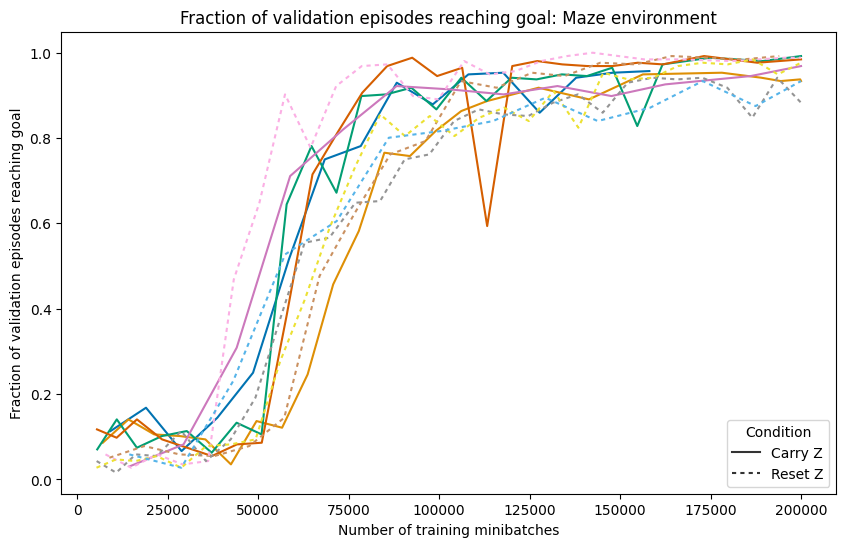}
      \caption{Random maze environment}
    \end{subfigure}
    \caption{Validation of goal-navigation performance: Fraction of episodes in which the goal is reached. The HRM-Agent model is able to reach the goal in approximately 99\% of all episodes in both environments. 5 models were trained with the recurrent state $z$ ``carried forward'' between each environment step. This allows the model to reuse and continue its current plan. 5 models were trained with recurrent state $z$ ``reset'' to a constant random vector at the start of each environment step. Each model training run is presented as a separate series.}
  \label{fig:training}
\end{figure*}

To validate that the model can learn to navigate successfully, figure \ref{fig:training} captures the fraction of episodes in which the agent reaches the goal. After sufficient training, the agent reaches the goal in approximately 99\% of episodes in both environments. We believe it is likely that a model with larger capacity (i.e. larger latent dimension size) and additional training time could reach 100\%, but considered these results sufficient to demonstrate effective navigation (to achieve more than 87.5\% success in the four-rooms environment with doors, the agent must be able to navigate through three doorways consecutively). Success in the random maze environment demonstrates a generalized ability to plan paths to the goal in unseen mazes.

\begin{figure*}[t!]
  \centering
    \begin{subfigure}[t]{0.49\textwidth}
      \centering
      \includegraphics[width=1.0\textwidth]{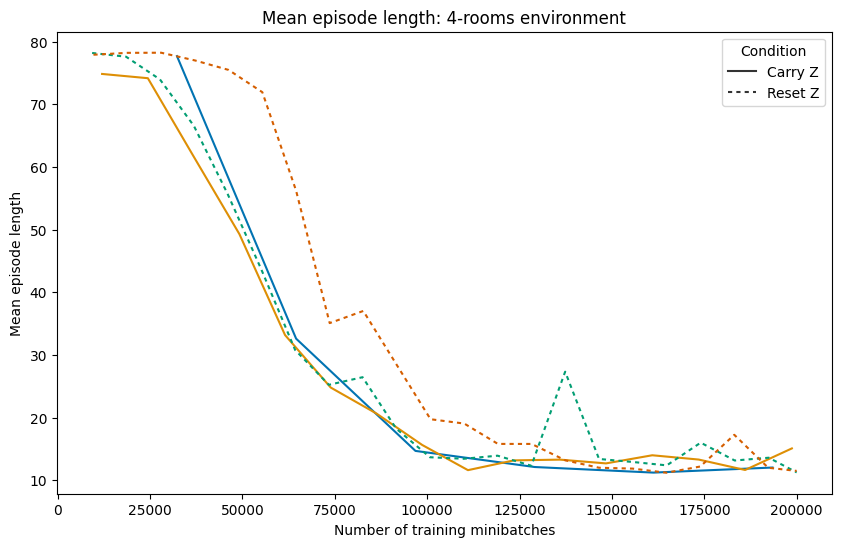}
      \caption{Four-rooms environment}
    \end{subfigure}
    \begin{subfigure}[t]{0.49\textwidth}
      \centering
      \includegraphics[width=1.0\textwidth]{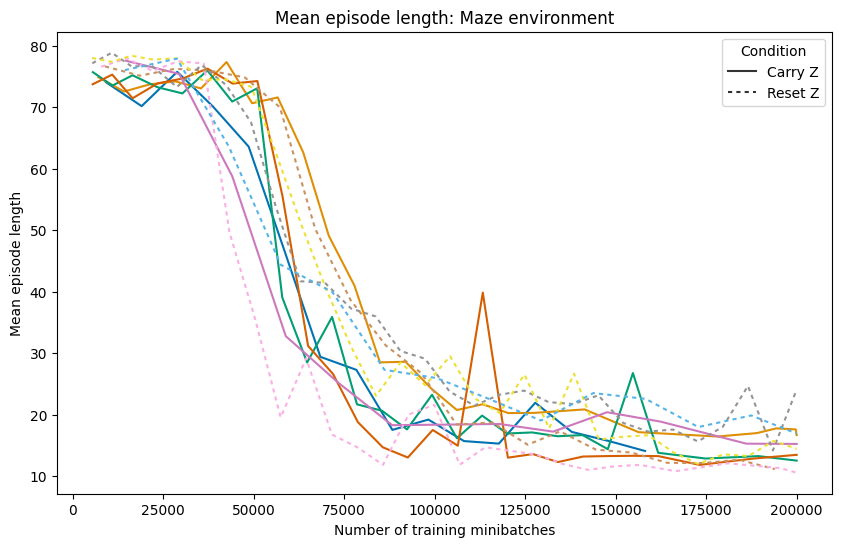}
      \caption{Random maze environment}
    \end{subfigure}
    \caption{Validation of efficient path planning: Mean episode length. Since episodes end when the goal is reached, short episodes indicate efficient paths. The theoretical optimum average path length in both environments is approximately 10 steps, when considering the effects of the random doors. The HRM-Agent model achieves efficient paths in both environments. Each series depicts the performance of a separate model training run (there are insufficient runs for a distributional plot).}
  \label{fig:episode-length}
\end{figure*}

Figure \ref{fig:episode-length} demonstrates that the HRM-Agent has learned to navigate \emph{efficiently}. After sufficient training, the mean episode length approaches the theoretical optimum mean path length in both environments (approximately 10 steps\footnote{In the original four-rooms environment the optimal mean path-length is approximately 6 steps, but the addition of closed doors requires the agent to either wait or take longer detours in roughly 37.5\% of episodes.}).

In the dynamic random maze environment we performed 5 training runs of each variant. In 4 of 5 training runs the ``Carry Z'' reaches high goal-achievement and efficient path lengths faster than the ``Reset Z'' variant.

\subsection{Analysis of recurrent convergence dynamics}
This section provides results which explore the behaviour of the recurrent process for the two model variants, ``Carry Z'' and ``Reset Z''. We hypothesized that the recurrent state $z$ contains part or all of the planned path to the goal, even though the agent only moves one square at a time. This is because in both environments the best \emph{next} action depends on knowledge of a viable path to the distant goal.

If this hypothesis is true we can make some predictions about the behaviour of $z$ between environment time-steps and during the recurrent convergence process. We analyzed four conditions which occur during validation episodes (i.e. evaluating the trained model without learning):

\begin{itemize}
\item Carry Z; environment \textbf{has} changed since the previous time-step (i.e. the status of door[s] have changed, potentially invalidating the previous plan)
\item Carry Z; environment has \textbf{not} changed
\item Reset Z (i.e. do not copy the initial values of $z_H$ and $z_L$ from their final values in the previous time-step); environment \textbf{has} changed
\item Reset Z; environment has \textbf{not} changed
\end{itemize}

\begin{figure}[htbp]
  \centering
  \includegraphics[width=1.0\textwidth]{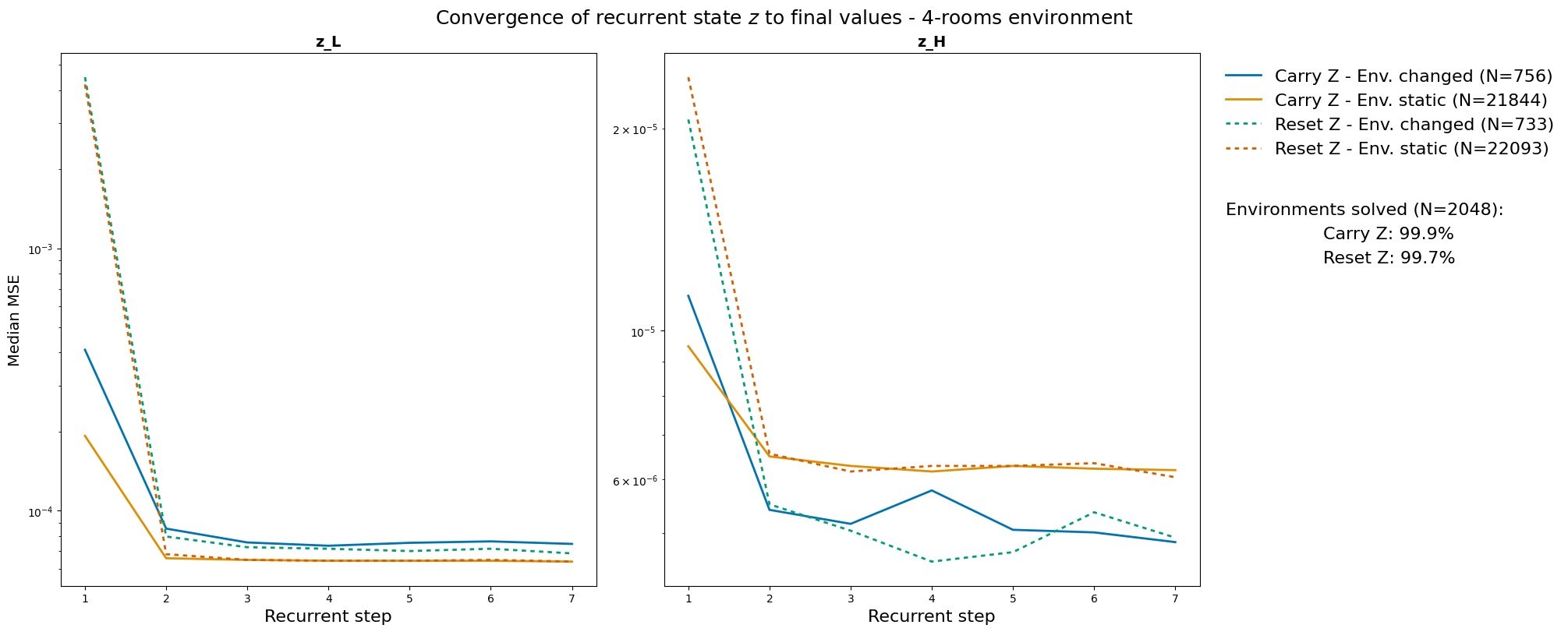}
  \caption{Analysis of the convergence of the recurrent latent state $z$ to its final values in the four-rooms environment. The left plot shows $z_L$ and the right plot $z_H$. Each series is shows the median mean-square-error (MSE) between $z^i$ at recurrent time step $i$ and the final $z^{NT}$, drawn from $N$ environment time-steps collected from many validation episodes. Solid lines depict model variants trained and evaluated with the ``Carry Z'' initialization of $z^0$. Dashed lines depict the ``Reset Z'' variant. The peak at recurrent step 4 in the $z_H$ plot appears to be a consistent feature of the dynamics in this environment; note that the iteration of both H and L modules are unrolled into a single axis, and this step displays the result of the 2nd update of the H module.}
    \label{fig:convergence-4-rooms}
\end{figure}

\begin{figure}[htbp]
  \centering
  \includegraphics[width=1.0\textwidth]{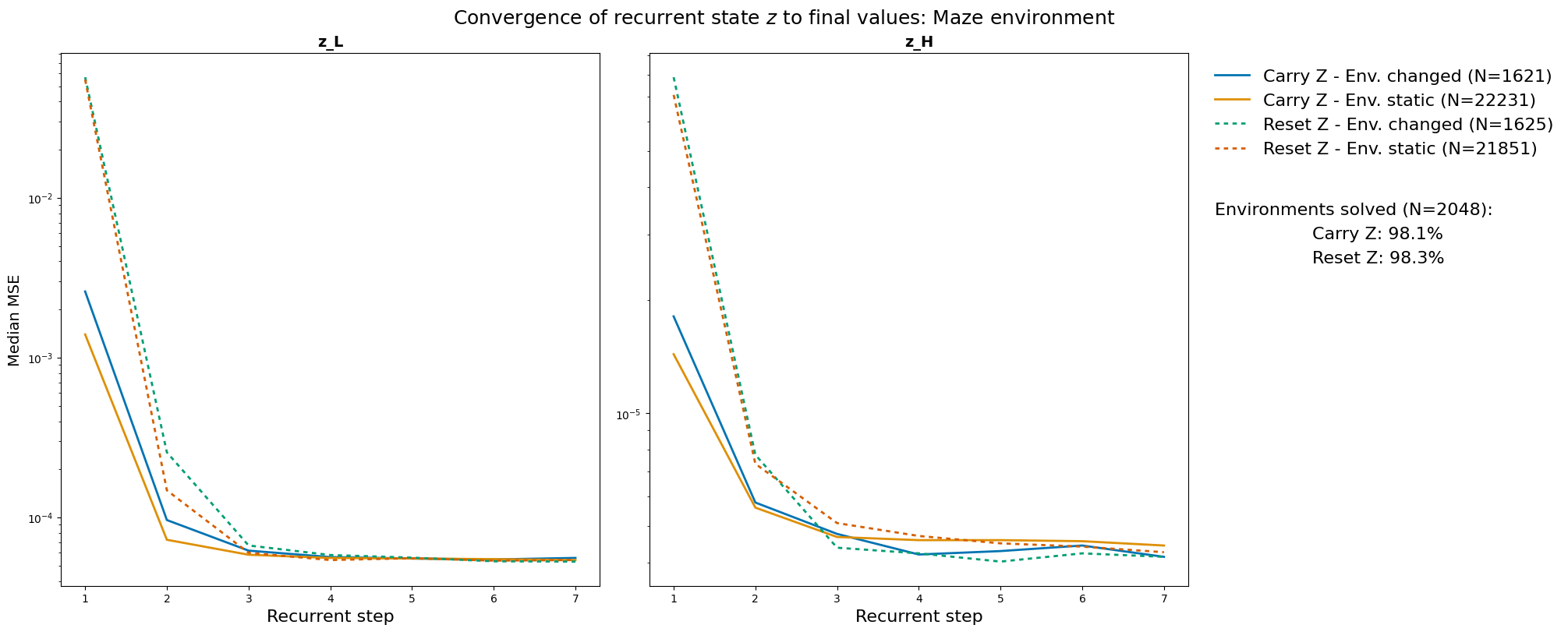}
  \caption{Analysis of the convergence of the recurrent latent state $z$ to its final values, in the dynamic random maze environment. See figure \ref{fig:convergence-4-rooms} for a description of the plot contents. The ``Carry Z'' model variant shows that carrying forward the previous latent state yields faster convergence, and that environmental change increases the dissimilarity of the initial and final latent state, which would be expected if the latent state describes a path to the goal.}
    \label{fig:convergence-maze}
\end{figure}

In each condition, we measure the mean-square-error (MSE) between $z^i_H$ at recurrent time step $i$ and the final $z^{NT}_H$ (using notation from the original HRM paper). The $z^1_H$ measured at the first recurrent time-step is the value \emph{after} the first forward pass, not the initial value. The same calculation is performed for $z_L$. These statistics are collated over $N$ episodes, and the median values are taken over all recorded episodes for each recurrent time-step $i$. This gives the plots shown in figures \ref{fig:convergence-4-rooms} and \ref{fig:convergence-maze}. These depict the convergence process by comparing the evolving recurrent state to its final values. By definition, the last comparison will always have an ``error'' of zero and is therefore not displayed.

Before examining these plots, consider the following:

\begin{itemize}
\item If the environment changes in a way which requires a different plan or path, the initial $z$ should be very different to the final $z$.
\item If the environment changes in a way which does not change the plan, the initial $z$ should be closer to the final $z$.
\item Since the four defined conditions do not distinguish between \emph{material} changes to the environment and irrelevant ones, any measured effect will be diluted. Material changes are closures of doors which block the shortest path to the goal.
\item In the ``Reset Z'' variant, $z_H$ and $z_L$ are initialized to a random vector and we predict a large initial change from this value.
\item In the ``Carry Z'' variant, if the model benefits from knowledge of its existing plan, convergence should be quicker and the distance from the initial to the final values of $z$ should be smaller, suggesting that the final ``plan'' is similar to the plan from the previous environment time-step.
\end{itemize}


The statements above broadly agree with the results displayed in figures \ref{fig:convergence-4-rooms} and \ref{fig:convergence-maze}. In both figures the distance between $z^1$ and $z^{NT}$ is smaller for the ``Carry Z'' variant. In both environments, for the ``Carry Z'' variant, the initial distance is greater when the environment has changed. In the Random Maze environment, the recurrent process converges faster with ``Carry Z'', both when the environment has changed, and when it has not \ref{fig:convergence-maze}.

\begin{figure}[htbp]
  \centering
  \includegraphics[width=1.0\textwidth]{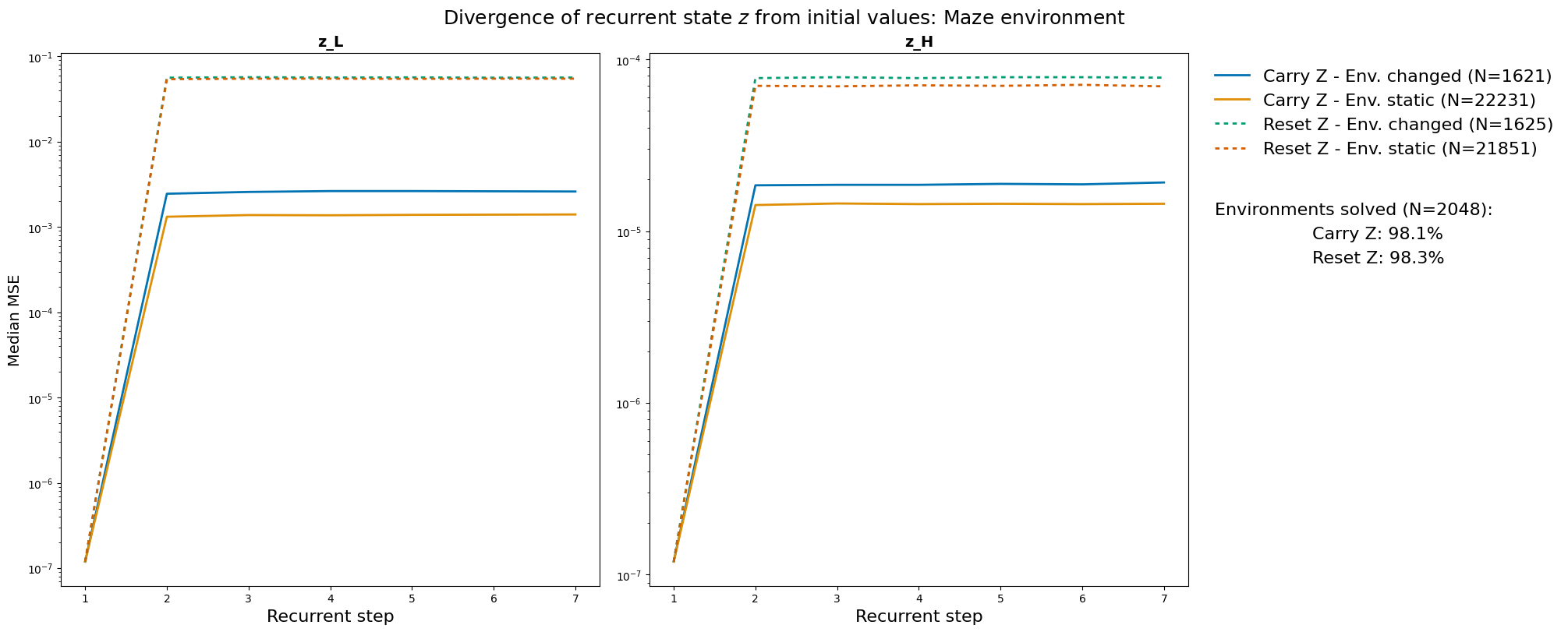}
  \caption{Analysis of the \textbf{divergence} of the recurrent latent state $z$ from the output of the first forward pass $z^1$ to its final values, in the dynamic random maze environment. See figure \ref{fig:convergence-4-rooms} for a description of how the plot series were created. The final latent state $z^{NT}$ is much closer to the initial latent state $z^1$ when the input to the initial forward pass is carried forward from the previous environment time-step. This suggests the resultant paths are also more consistent in this condition, which can be important for execution of complex plans in dynamic environments.}
    \label{fig:divergence-maze}
\end{figure}

These results support the belief that the agent is constructing a representation of the path to the goal in its recurrent latent state (i.e. is reasoning), before generating the next action from this plan. If the ``Carry Z'' option is used, the Agent shows signs that it can reuse earlier computation. The divergence plot (figure \ref{fig:divergence-maze}) also provides evidence that the ``Carry Z'' variant produces actions which are more \emph{consistent} with the plan from the previous environment time-step. Carrying forward the latent state reduces the divergence by a greater magnitude than whether the environment has changed or not.

\section{Discussion}
We demonstrated a basic proof of concept (PoC) of a HRM model variant trained exclusively through reinforcement learning. 

Prior to completing experiments it was not clear whether ``carrying forward'' $z_H$ and $z_L$ would help or harm model training, but we believe this is essential to be able to use a recurrent reasoning model in a dynamic environment efficiently and with continuity of intent, without treating each new moment as an entirely new problem.

Although both environments are simple maze problems, observed performance requires deliberate planning rather than simple heuristics. The results also show that ``carrying forward'' the recurrent state does allow the model to start from a more similar initial position and converge more rapidly, efficiently reusing computation from previous environment time-steps.

But faster convergence in a reasoning process can be harmful. The effects on model training could be especially pathological, especially during the early stages of learning, when it might limit the entropy of the latent state. If a lack of entropy is observed in more complex problems, this issue might be ameliorated by only ``carrying forward'' $z$ with probability $p$ (similar to dropout \citep{srivastava2014dropout}).

Despite these concerns, the results we have obtained so far indicate that maintaining the recurrent state is actually helpful both in terms of recurrent time-to-convergence, and overall training time (measured in environment time-steps), although evaluation on more complex tasks would be important to fully substantiate this.

\subsection{Future work}
To evaluate a HRM-Agent in more complex environments a more sophisticated model will be necessary. We intend to develop an on-policy, A2C \citep{mnih2016asynchronous} model variant. We will also restore the Adaptive Computation Time (ACT) feature, and verify that the model can learn to optimize its ``thinking time'' while simultaneously learning to reason.

The recent ``Tiny Recursive Model'' derivative of HRM \citep{jolicoeurmartineau2025less} is appealing due to its combination of relative simplicity and good reasoning performance. We will investigate simplifying the next model accordingly.

The NetHack Learning Environment is open-ended and can easily be extended to include more complex dynamics, such as monsters and lockable doors. The MiniHack library also supports an agent-centred observation mode, which will allow us to validate training of a HRM-Agent in a partially-observable setting. The transformer input embedding stream can easily be updated to allow attention over a moving window of recent observations.

Beyond those enhancements, we would like to explore a recurrent reasoning model which can perform continual and few-shot learning (CFSL) in an online, streaming learning framework. While this is extremely challenging, recent reports suggest significant progress training RL models in these conditions \citep{elsayed2024streaming}.


\begin{ack}
We would like to acknowledge and thank \href{https://cerenaut.ai}{Cerenaut.ai} and the \href{https://wba-initiative.org/en/}{Whole Brain Architecture Initiative (WBAI)} for their support, which helped to fund compute resources used in this and other recent work. In particular, this work was supported by WBAI's Incentive Award (2024).
\end{ack}


\medskip

{
\small

\bibliographystyle{plainnat}
\bibliography{references}




}


\appendix

\section{Hyperparameters}
Table \ref{table:params} lists the values which were used in all experiments. The total trainable parameter count is 541,186. All experiments were conducted on a single NVIDIA L4 GPU.

\begin{table}[htbp]
  \caption{Hyperparameters}
  \label{table:params}
  \centering
  \begin{tabular}{ll}
    \toprule
    Description     & Value \\
    \midrule
    Observation size tokens & 121     \\
    Action space size & 4     \\
    Max. episode length & 80     \\
    Training batch size & 256     \\
    Collector workers & 5     \\
    Collector size & 1,048,576     \\
    Collector update interval (in batches) & 8     \\
    
    HRM hidden size & 64     \\
    HRM recurrent max. steps & 8     \\
    HRM H cycles & 2     \\
    HRM L cycles & 2     \\
    HRM H layers & 4     \\
    HRM L layers & 4     \\
    HRM transformer heads & 2     \\
    HRM expansion & 4     \\
    Model optimizer learning rate & 1e-4     \\
    Model optimizer beta 1 & 0.9     \\
    Model optimizer beta 2 & 0.95     \\
    Model optimizer weight decay & 0     \\
    Positional embedding type & RoPE     \\
    Embedding optimizer learning rate & 1e-2     \\
    Embedding optimizer weight decay & 0.1     \\
    DQN discount factor & 0.95     \\
    DQN epsilon initial & 1     \\
    DQN epsilon final & 0.15     \\
    DQN epsilon decay period & 300,000     \\
    DQN use target network & True     \\
    DQN target network delay factor & 0.999     \\
    \bottomrule
  \end{tabular}
\end{table}


\end{document}